\documentclass[journal]{IEEEtran}
\usepackage{cite}

\ifCLASSINFOpdf
  \usepackage[pdftex]{graphicx}
\else
  \usepackage[dvips]{graphicx}
\fi
\ifCLASSOPTIONcompsoc
  \usepackage[caption=false,font=normalsize,labelfont=sf,textfont=sf]{subfig}
\else
  \usepackage[caption=false,font=footnotesize]{subfig}
\fi
\usepackage{url}

\usepackage{hyperref}

\usepackage[many]{tcolorbox}

\NewTColorBox{NewBox}{ s O{!htbp} }{%
  floatplacement={#2},
  IfBooleanTF={#1}{float*,width=\textwidth}{float},
  colframe=blue!50!black,colback=blue!10!white,
  title=MAPF Solvers
  }

\begin{document}
%
\title{Overview: A Hierarchical Framework \\ for Plan Generation and Execution
  \\ in Multi-Robot Systems}
%
%
%

\author{Hang~Ma,
        Wolfgang~H\"onig,
        Liron~Cohen,
        Tansel~Uras,
        Hong~Xu,
        T.~K.~Satish~Kumar,
        Nora~Ayanian,
        and Sven~Koenig
        \\University of Southern California}

%
%

\markboth{IEEE Intelligent Systems}%
{Shell \MakeLowercase{\textit{et al.}}: Bare Demo of IEEEtran.cls for IEEE Journals}
%



\maketitle

\begin{abstract}
We present an overview of a hierarchical framework for coordinating task-level
and motion-level operations in multi-robot systems. Our framework is based on
the idea of using simple temporal networks to simultaneously reason about (a)
precedence/causal constraints required for task-level coordination and (b)
simple temporal constraints required to take some kinematic constraints of
robots into account. In the plan-generation phase, our framework provides a
computationally scalable method for generating plans that achieve high-level
tasks for groups of robots and take some of their kinematic constraints into
account. In the plan-execution phase, our framework provides a method for
absorbing an imperfect plan execution to avoid time-consuming re-planning in
many cases. We use the multi-robot path-planning problem as a case study to
present the key ideas behind our framework for the long-term autonomy of
multi-robot systems.
\end{abstract}

\begin{IEEEkeywords}
Multi-robot planning, multi-robot systems, path planning, plan execution.
\end{IEEEkeywords}

%
\IEEEpeerreviewmaketitle

\section{Introduction}

%
%
%
%

\IEEEPARstart{T}{}he problem of coordinating task-level and motion-level
operations for multi-robot systems arises in many real-world scenarios. A
simple example is an automated-warehouse system where heavy robots move
inventory pods in a space inhabited by humans. The robots may have to avoid
close proximity to humans and each other; they may have to compete for
resources with each other; and, yet, they have to work toward a common
objective~\cite{kiva}. Another example is airport surface operations where
towing vehicles autonomously navigate to aircraft and tow them to their
destinations~\cite{airporttug16}. This task-level coordination has to be done
in conjunction with the motion-level coordination of action primitives so that
each robot has a kinematically feasible plan.

The coordination of task-level and motion-level operations for multi-robot
systems requires a large search space. Current technologies are inadequate for
addressing the complexity of the problem, which becomes even worse since we
have to take imperfections in plan execution into account. For example,
exogenous events may not be included in the domain model. Even if they are,
they can often be modeled only probabilistically~\cite{MaAAAI17}.

\begin{figure*}
\center
  \includegraphics[width=\textwidth]{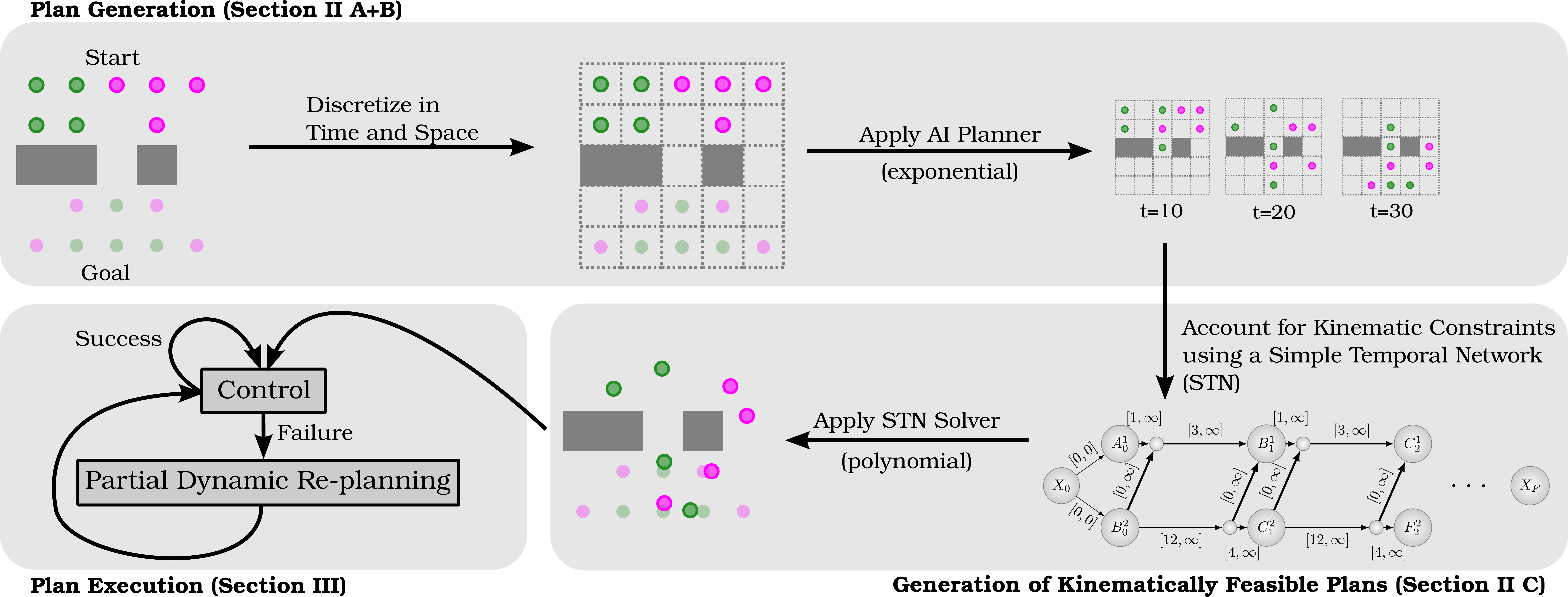}
  \caption{Architecture of our hierarchical framework. First, we discretize
    the continuous MAPF problem in time and space and use an AI planner to
    solve the resulting NP-hard problem. Then, we solve the STN for the
    resulting discrete MAPF plan in polynomial time to generate a
    kinematically feasible plan that provides guaranteed safety distances
    among robots under the assumption of perfect plan execution.  Control uses
    specialized robot controllers during plan execution to exploit the slack
    in the plan to try to absorb any imperfect plan execution. If this does
    not work, partial dynamic re-planning re-solves a suitably modified STN in
    polynomial time. Only if this does not work either, partial dynamic
    re-planning re-solves a suitably modified MAPF problem more slowly.}
    \label{fig:architecture}
\end{figure*}

In this article, we present an overview of our hierarchical framework for the
long-term autonomy of multi-robot systems. Our framework combines techniques
from automated artificial intelligence (AI) planning, temporal reasoning and
robotics. Figure~\ref{fig:architecture} shows its architecture for a small
example.

The plan-generation phase uses a state-of-the-art AI
planner~\cite{CohenUK16,MaAAMAS16} for causal reasoning about the task-level
actions of the robots, independent of their kinematic constraints to achieve
scalability. It then identifies the dependencies between the preconditions and
effects of the actions in the generated plan and compiles them into a temporal
plan graph (TPG), that encodes their partial temporal order. Finally, it
annotates the TPG with quantitative information that captures some kinematic
constraints associated with executing the actions. This converts the TPG into
a simple temporal network (STN) from which a plan (including its execution
schedule) can be generated in polynomial time that takes some of the kinematic
constraints of the robots into account (for simplicity called a kinematically
feasible plan in the following), namely by exploiting the slack in the
STN. The term ``slack'' refers to the existence of an entire class of plans
consistent with the STN, allowing us to narrow down the class of plans to a
single kinematically feasible plan.  A similar notion of slack is well studied
for STNs in general in the temporal-reasoning community.

The plan-execution phase also exploits the slack in the STN, namely for
absorbing any imperfect plan execution to avoid time-consuming re-planning in
many cases.

We use a multi-robot path-planning problem as a case study to present the key
ideas behind our framework and demonstrate it both in simulation and on real
robots.

\section{Plan Generation}

We use a state-of-the-art AI planner for reasoning about the causal
interactions among actions. In the multi-agent path-finding (MAPF) problem,
which is well studied in AI, robotics and theoretical computer science, the
causal interactions are studied oblivious to the kinematic constraints of the
robots. We are given a graph with vertices (that correspond to locations) and
unit-length edges between them. Each edge connects two different vertices and
corresponds to a narrow passageway between the corresponding locations in
which robots cannot pass each other. Given a set of robots with assigned start
vertices and targets (goal vertices), we have to find collision-free paths for
the robots from their start vertices to their targets (where the robots
remain) that minimize the makespan (or some other measure of the cost, such as
the flowtime). At each timestep, a robot can either wait at its current vertex
or traverse a single edge. Two robots collide when they are at the same vertex
at the same timestep or traverse the same edge at the same timestep in
opposite directions.

The MAPF problem is NP-hard to solve optimally or bounded sub-optimally since
it is NP-hard to approximate within any constant factor less than 4/3
\cite{MaAAAI16}, called the sub-optimality guarantee. Yet, powerful MAPF
planners have recently been developed in the AI community that can find
(optimal or bounded sub-optimal) collision-free plans for hundreds of robots at
the cost of ignoring the kinematic constraints of real robots
\cite{MaAAMAS16,CohenUK16,MaAAAI17,MaAAMAS17}. We report on two of our own
contributions to such MAPF planners below.

\subsection{Consistency and Predictability of Motion}

For many real-world multi-robot systems, the consistency and predictability of
robot motions is important (especially in work spaces shared by humans and
robots), which is not taken into account by existing MAPF planners. We have
shown that we can adapt AI planners, such as the bounded-sub-optimal MAPF
planner Enhanced Conflict-Based Search (ECBS)~\cite{ECBS}, to generate paths
that include edges from a user-provided set of edges (called highways)
whenever the sub-optimality guarantee allows it, which makes the robot motions
more consistent and thus predictable. The highways can be an arbitrary set of
edges and thus be chosen to suit the humans. For example, highways need to be
created only in the part of the environment where the consistency of robot
motions is important. Furthermore, highways provide only suggestions but not
restrictions. Poorly chosen highways do not make a MAPF instance unsolvable
although they can make the MAPF planner less efficient. On the other hand,
well chosen highways typically speed up the MAPF planner because they avoid
front-to-front collisions between robots that travel in opposite directions.

\begin{figure}
\center
  \includegraphics[width=\columnwidth]{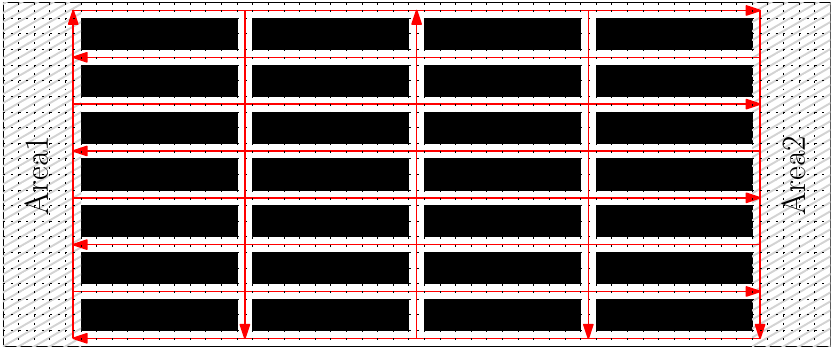}
  \caption{Environment of a simulated automated-warehouse system where robots
    need to swap sides from Area1 to Area2 and vice versa. The red arrows show
    user-suggested edges to traverse (called highways). Highways make the
    resulting plan more predictable and speed up planning.}\label{fig:arrows}
\end{figure}

Our version of the ECBS planner with highways either inflates the heuristic
values or the edge costs non-uniformly in a way that encourages path finding
to return paths that include the edges of the highways
\cite{DBLP:conf/socs/CohenUK15}. For example, we can place highways in an
automated-warehouse system along the narrow passageways between the storage
locations as shown by the red arrows in Figure~\ref{fig:arrows}. We have also
developed an approach for learning good highways automatically
\cite{CohenUK16}. It is based on the insight that solving the MAPF problem
optimally is NP-hard but computing the minimum-cost paths for all robots
independently is fast, by employing a graphical model that uses the
information in these paths heuristically to generate good highways
automatically.

\subsection{Target Assignment and Path Finding}

\begin{figure}
\center
  \includegraphics[height=50pt]{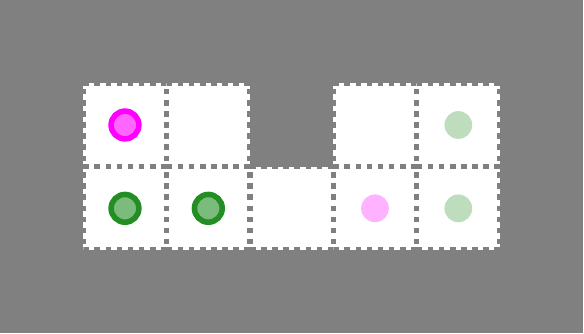} \includegraphics[height=50pt]{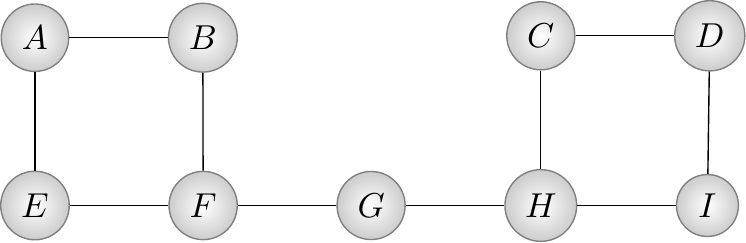}
  \caption{Left: TAPF instance with two teams: Team 1 (in pink) and Team 2 (in
    green). The circles on the left are robots. The circles in light colors on
    the right are targets given to the team of the same color. Right: Graph
    representation of the TAPF instance. Team 1 consists of a single robot
    with start vertex $A$ and target $H$. Team 2 consists of two robots with
    start vertices $E$ and $F$, respectively, and targets $D$ and
    $I$.}\label{fig:TAPF}
\end{figure}

For the MAPF problem, the assignments of robots to targets are pre-determined,
and robots are thus not exchangeable. In practice, however, the assignments of
robots to targets are often not predetermined. For example, consider two
robots in an automated-warehouse system that have to deliver two inventory
pods to the same packing station. It does not matter which robot arrives first
at the packing station, and their places in the arrival queue of the packing
station are thus not pre-determined.  We therefore define the combined target
assignment and path finding (TAPF) problem for teams of robots as a
combination of the target-assignment and path-finding problems.  The TAPF
problem is a generalization of the MAPF problem where the robots are
partitioned into equivalence classes (called teams). Each team is given the
same number of unique targets as there are robots in the team. We have to
assign the robots to the targets and find collision-free paths for the robots
from their start vertices to their targets in a way such that each robot moves
to exactly one target given to its team, all targets are visited and the
makespan is minimized. Any robot in a team can be assigned to any target of
the team, and robots in the same team are thus exchangeable. However, robots
in different teams are not exchangeable. Figure~\ref{fig:TAPF} shows a TAPF
instance with two teams of robots.

The TAPF problem is NP-hard to solve optimally or bounded sub-optimally for
more than one team \cite{MaAAAI16}. TAPF planners have two advantages over
MAPF planners: 1. Optimal TAPF plans often have smaller makespans than optimal
MAPF plans for TAPF instances since optimal TAPF plans optimize the
assignments of robots to targets. 2. State-of-the-art TAPF planners compute
collision-free paths for all robots on a team very fast and thus often scale
to a larger number of robots than state-of-the-art MAPF planners. We have
developed the optimal TAPF planner Conflict-Based Min-Cost Flow
(CBM)~\cite{MaAAMAS16}, that combines heuristic search-based MAPF planners
\cite{DBLP:journals/ai/SharonSFS15} and flow-based MAPF planners
\cite{YuLav13STAR} and scales to TAPF instances with dozens of teams and
hundreds of robots.

\subsection{Generation of Kinematically Feasible Plans}

MAPF/TAPF planners generate plans using idealized models that do not take the
kinematic constraints of actual robots into account. For example, they gain
efficiency by not taking velocity constraints into account and instead
assuming that all robots always move with the same nominal speed in perfect
synchronization with each other. However, it is communication-intensive for
robots to remain perfectly synchronized as they follow their paths, and their
individual progress will thus typically deviate from the plan. Two robots can
collide, for example, if one robot already moves at large speed while another
robot accelerates from standstill. Slowing down all robots results in large
makespans and is thus undesirable.

\begin{figure}
\small
\centering
\footnotesize
\begin{tabular}{c|cccc}
Robot &$t=1$ &$t=2$ &$t=3$ &$t=4$\\
\hline
$1$ (in Team 1) 	&$A \to B$	&$B \to F$	&$F \to G$	&$G \to H$ \\
$2$ (in Team 2) 	&$E \to F$	&$F \to G$	&$G \to H$	&$H \to I$ \\
$3$ (in Team 2) 	&$F \to G$	&$G\to H$	&$H \to C$	&$C \to D$ \\
\end{tabular}
\includegraphics[width=\columnwidth]{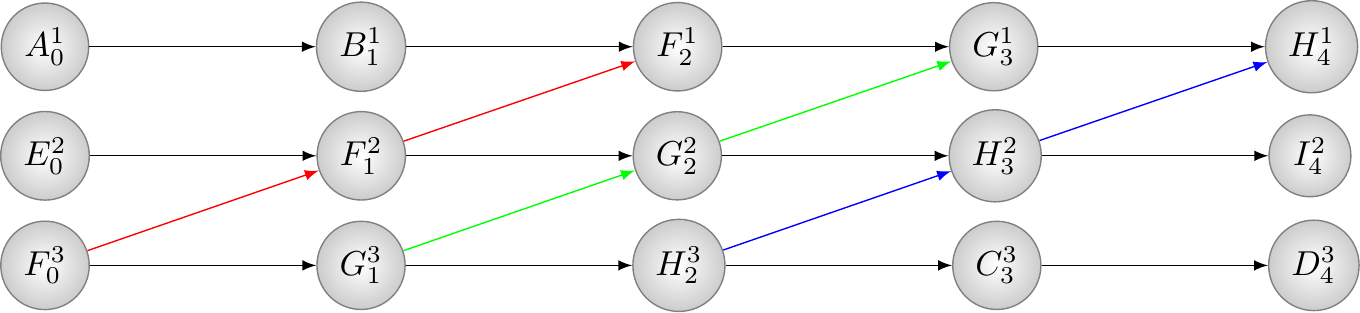}\\
\vspace{0.1cm}
\includegraphics[width=\columnwidth]{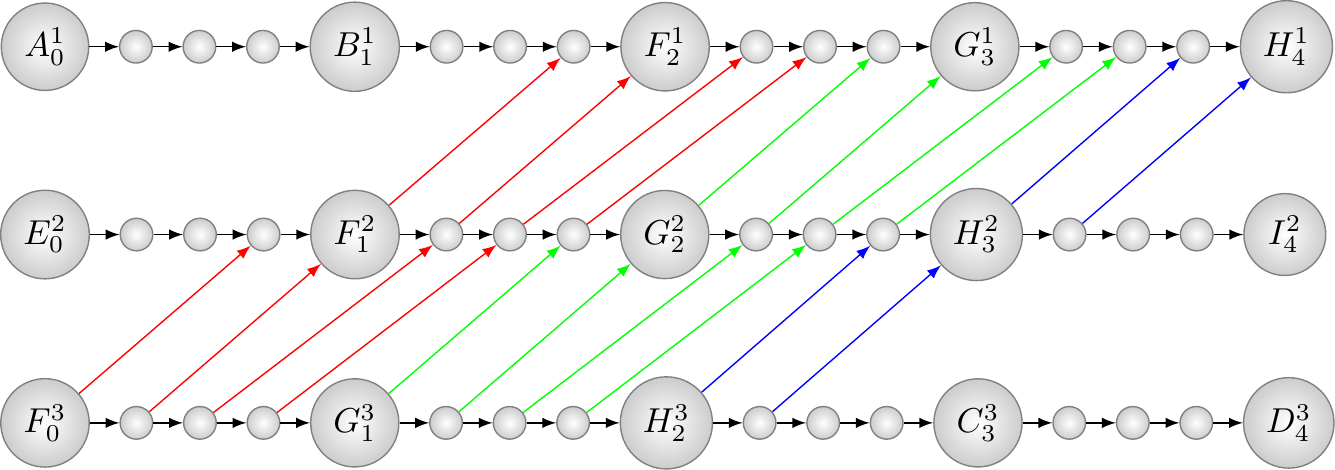}
\caption{Top: TAPF plan produced by the optimal TAPF planner CBM for the TAPF
  instance in Figure \ref{fig:TAPF}. Middle: TPG for the TAPF plan. Each node
  $l_i^j$ in the TPG represents the event ``robot $j$ arrives at vertex $l$''
  at timestep $i$. The arcs indicate temporal precedences between
  events. Bottom: Augmented TPG.}
\label{fig:STN}
\end{figure}

We have thus developed MAPF-POST, a novel approach that makes use of a simple
temporal network (STN)~\cite{Dechter1991} to postprocess a MAPF/TAPF plan in
polynomial time and create a kinematically feasible
plan~\cite{HoenigICAPS16,HoenigIROS16}. MAPF-POST utilizes information about
the edge lengths and maximum translational and rotational velocities of the
robots to translate the plan into a temporal plan graph (TPG) and augment the
TPG with additional nodes that guarantee safety distances among the
robots. Figure~\ref{fig:STN} shows an example. Then, it translates the
augmented TPG into an STN by associating bounds with arcs in the augmented TPG
that express non-uniform edge lengths or velocity limits (due to kinematic
constraints of the robots or safety concerns). It then obtains an execution
schedule from the STN by minimizing the makespan or maximizing the safety
distance via graph-based optimization or linear programming. The execution
schedule specifies when each robot should arrive in each location of the plan
(called arrival times). The kinematically feasible plan is a list of locations
(that specify way-points for the robots) with their associated arrival times.
See \cite{HoenigIROS16} for more details.

\section{Plan Execution}

The robots will likely not be able to follow the execution schedule perfectly,
resulting in plan deviations. For example, our planner takes velocity
constraints into account but does not capture higher-order kinematic
constraints, such as acceleration limits. Also, robots might be forced to slow
down due to unforeseen exogenous events, such as floors becoming slippery due
to water spills. In such cases, the plan has to be adjusted quickly during
plan execution.

Frequent re-planning could address these plan deviations but is time-consuming
(and thus impractical) due to the NP-hardness of the MAPF/TAPF problem.
Instead, control uses specialized robot controllers to exploit the slack in
the plan to try to absorb any imperfect plan execution. If this does not work,
partial dynamic re-planning re-solves a suitably modified STN in polynomial
time. Only if this does not work either, partial dynamic re-planning re-solves
a suitably modified MAPF problem more slowly.

\subsection{Control}

A robot controller takes the current state and goal as input and computes the
motor output. For example, the state of a differential drive robot can be its
position and heading, and the motor output is the velocities of the two
wheels. The goal is the execution schedule, assuming a constant movement
velocity between two consecutive way-points (called the constant velocity
assumption).  Robots cannot execute such motion directly because they cannot
change their velocities instantaneously and might not be able to move
sideways. The actual safety distance during plan execution is thus often
smaller than the one predicted during planning, which is why we recommend to
maximize the safety distance during planning rather than the makespan. We use
robot controllers that try to minimize the effect of the above limitations.
For differential drive robots, we use the fact that turning in place is often
much faster than moving forward. Furthermore, we adjust the robot velocities
dynamically based on the time-to-go to reach the next way-point.  It is
especially important to monitor progress toward locations that correspond to
nodes whose slacks are small. Robots could be alerted of the importance of
reaching these bottleneck locations in a timely manner. Similar control
techniques can be used for other robots as well, such as drones, as long as no
aggressive maneuvers are required.

\subsection{Partial Dynamic Re-planning}

If control is insufficient to achieve the arrival times given in the execution
schedule, we adjust the arrival times by re-solving a suitably modified STN,
resulting in a new execution schedule. Only if this does not work either, we
re-solve a suitably modified MAPF problem, resulting in a new kinematically
feasible plan. If probabilistic models of delays and other deviations from the
nominal velocities are available, they could be used to determine the
probabilities that each location will be reached in a certain time interval
and trigger re-planning only if one or more of these probabilities become
small~\cite{HoenigICAPS16}.

\section{Experiments}
We have implemented our approach in C++ using the boost library for advanced
data structures, such as graphs. Experiments can be executed on three
abstraction levels, namely (a) an agent simulation, (b) a robot simulation and
(c) real robots:

\begin{itemize}

\item The agent simulation uses the constant velocity assumption and is
  fast. It can be used to verify the code and create useful statistics for the
  runtime, minimum distance between any two robots and average time until any
  robot reaches its target, among others. It can also be used for scalability
  experiments with hundreds of robots in cluttered environments.

\item The robot simulation adds realism because it uses a physics engine
  (instead of the constant velocity assumption) and realistic robot
  controllers for the simulated robots to follow the execution schedule. We
  use V-REP as robot simulation for differential drive robots, robots with
  omni-directional wheels, flying robots and spider-like robots.

\item Real robots are the ultimate testbed. We use a team of eight iRobot
  Create2 differential drive robots~\cite{HoenigIROS16}.

\end{itemize}

In the following, we discuss two example use cases on a 2.1 GHz Intel Core
i7-4600U laptop computer with 12 GB RAM. Each example is solved within 10
seconds of computation time and also shown in our supplemental video at
\url{http://idm-lab.org/project-p.html}.

\subsection{Automated Warehouse}

In the automated-warehouse use case, we model two robot teams. The first team
consists of ten KUKA youBot robots, which are robots with omni-directional
wheels capable of carrying (only) small boxes. The second team consists of two
Pioneer P3DX robots, which are differential-drive robots capable of carrying
(only) large boxes. The robots have to pick up small and large color-coded
boxes and bring them to a target of the same color. We split the task into two
parts.  First, each robot has to move to an appropriately sized box and pick
it up. Second, it has to move to a target of the same color. The first part is
a TAPF instance with two teams, one for each robot type. The second part is a
TAPF instance with four teams, one for each color.

\begin{figure}
\center
  \includegraphics[width=\columnwidth]{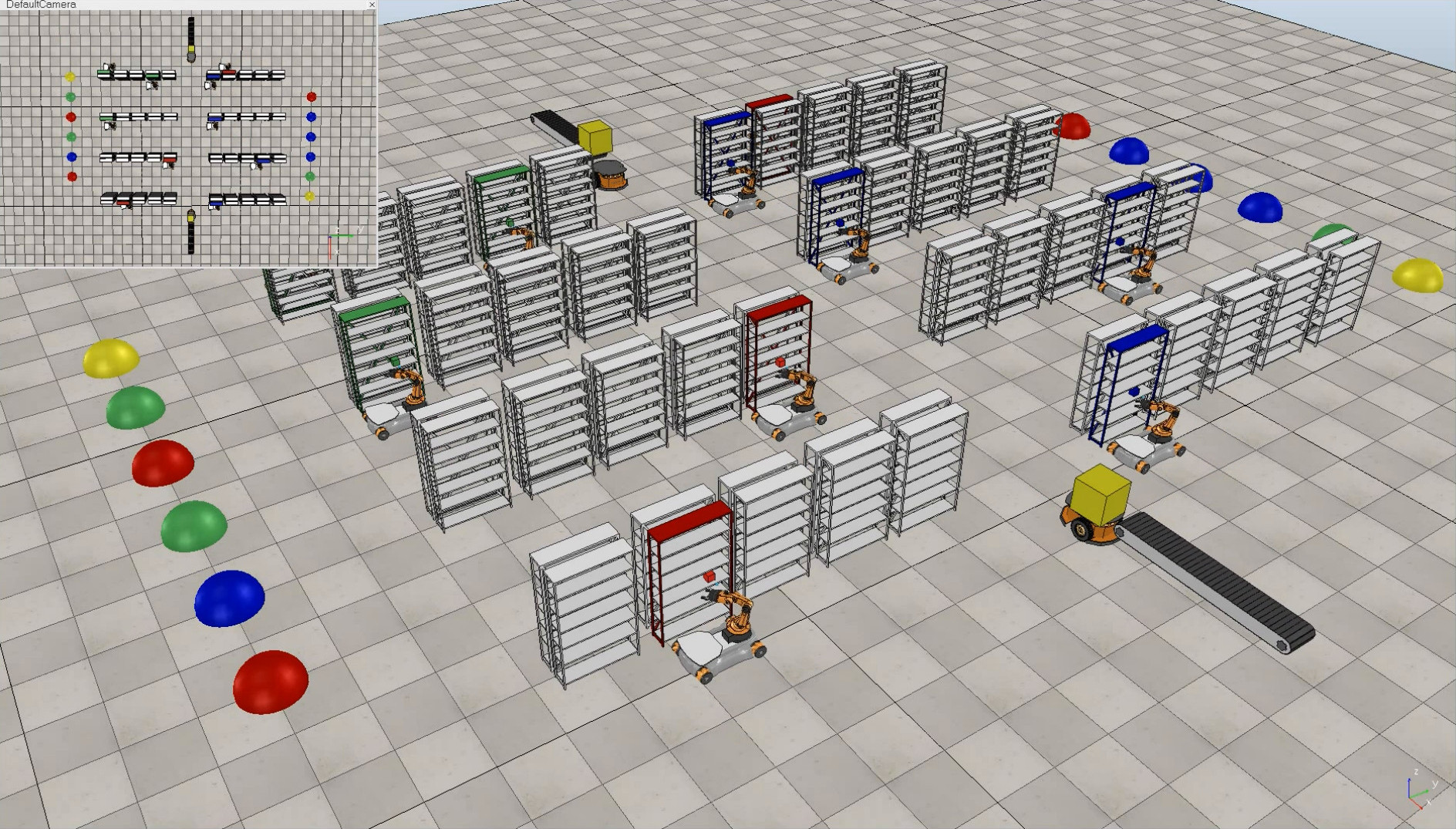}
  \caption{Simulated automated-warehouse environment. The in-set in the
    top-left corner shows an overhead view. The robots are at different
    pick-up locations and need to deliver the color-coded boxes to the left
    and right side, respectively.}\label{fig:exp:warehouse}
\end{figure}

We use the robot simulation on a 2D grid. Figure~\ref{fig:exp:warehouse} shows
a screen-shot after the first part has already been executed, and the robots
are at different pick-up locations. The KUKA robots use their grippers to pick
small boxes from shelves while the Pioneer robots receive the large boxes from
a conveyor belt. The robots then need to move to the targets on the left and
right side of the warehouse, respectively.

\subsection{Formation Changes}
Formations are useful for convoys, surveillance operations and artistic
shows. The task of switching from one formation to another, perhaps in a
cluttered environment, is a TAPF problem. In the formation-change use case, we
model a team of 32 identical quadcopters that start in a building with five
open doors. The robots have to spell the letters U -- S -- C outside the
building, which is a special TAPF instance where all robots are exchangeable
(also called an anonymous MAPF instance \cite{YuLav13STAR}).

\begin{figure}
\center
  \includegraphics[width=\columnwidth]{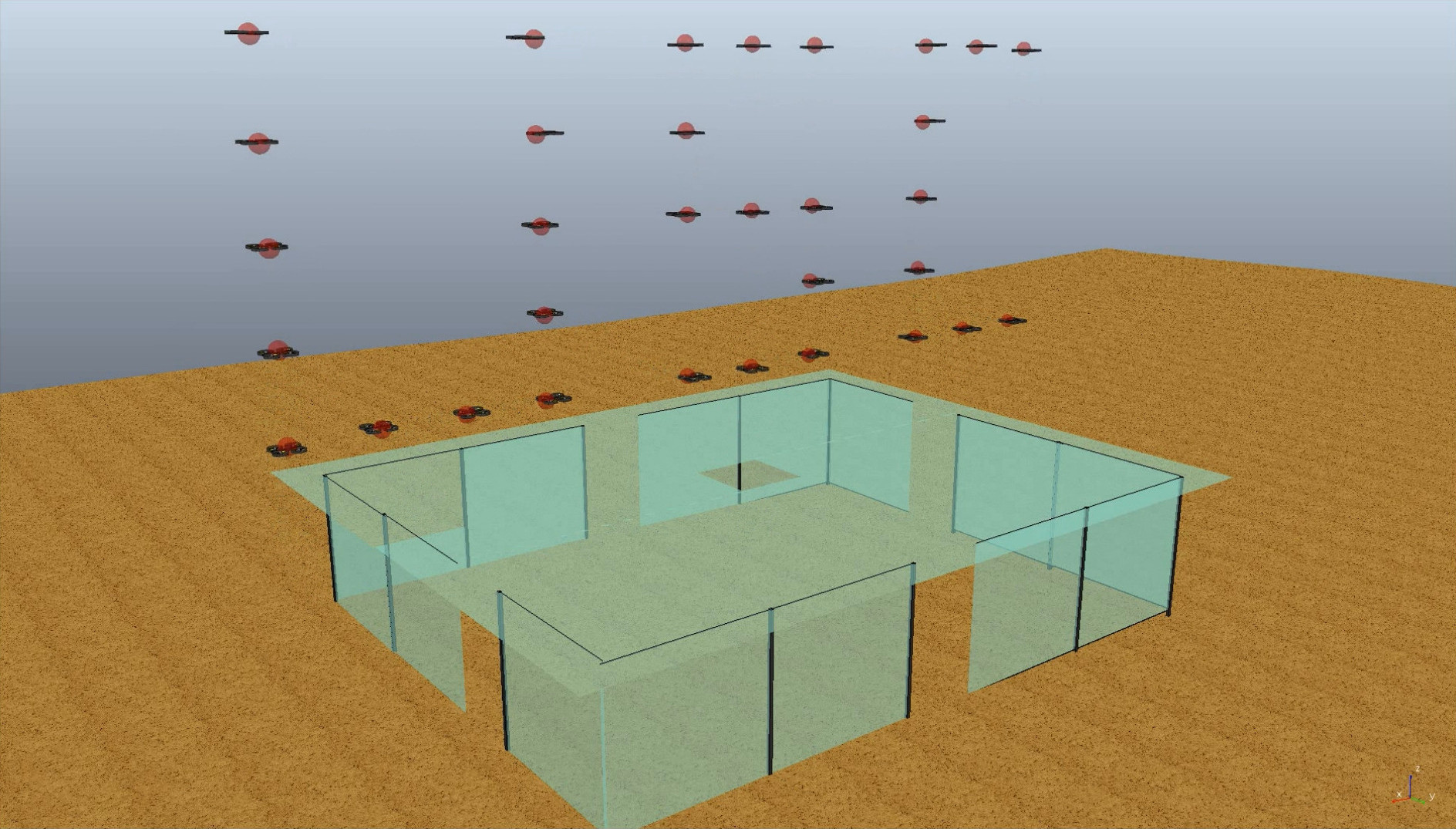}
  \caption{Simulated formation-change environment. 32 quadcopters start
    inside the glass building at the bottom of the picture and need to
    coordinate the usage of the four exit doors in order to create the
    depicted goal formation spelling the letters U -- S --
    C.}\label{fig:exp:formations}
\end{figure}

We use the robot simulation on a 3D grid. Figure~\ref{fig:exp:formations}
shows a screen-shot of the goal formation.

\section{Conclusions}

We presented an overview of our hierarchical framework for coordinating
task-level and motion-level operations in multi-robot systems using the
multi-robot path-planning problem as a case study. We use a state-of-the-art
AI planner for causal reasoning. The AI planner exploits the problem structure
to address the combinatorics of the multi-robot path-planning problem but is
oblivious to the kinematic constraints of the robots. We make the plan
kinematically feasible by identifying the causal dependencies among its
actions and embedding them in an STN. We then use the slack in the STN to
create a kinematically feasible plan and absorb any imperfect plan execution
to avoid time-consuming re-planning in many cases. For more information on our
research, see \url{http://idm-lab.org/project-p.html}.

\section*{Acknowledgments}

Our research was supported by ARL under grant number W911NF-14-D-0005, ONR
under grant numbers N00014-14-1-0734 and N00014-09-1-1031, NASA via Stinger
Ghaffarian Technologies and NSF under grant numbers 1409987 and 1319966. The
views and conclusions contained in this document are those of the authors and
should not be interpreted as representing the official policies, either
expressed or implied, of the sponsoring organizations, agencies or the
U.S. government.

\ifCLASSOPTIONcaptionsoff
  \newpage
\fi



%
\bibliographystyle{IEEEtran}
\bibliography{references}

%


\newpage
\begin{IEEEbiographynophoto}{Hang Ma}
is a Ph.D. student in the Department of Computer Science at the University of Southern California. Contact him at hangma@usc.edu.
\end{IEEEbiographynophoto}

\begin{IEEEbiographynophoto}{Wolfgang H\"onig}
is a Ph.D. student in the Department of Computer Science at the University of Southern California. Contact him at whoenig@usc.edu.
\end{IEEEbiographynophoto}

\begin{IEEEbiographynophoto}{Liron Cohen}
is a Ph.D. student in the Department of Computer Science at the University of Southern California. Contact him at lironcoh@usc.edu.
\end{IEEEbiographynophoto}

\begin{IEEEbiographynophoto}{Tansel Uras}
is a Ph.D. student in the Department of Computer Science at the University of Southern California. Contact him at turas@usc.edu.
\end{IEEEbiographynophoto}

\begin{IEEEbiographynophoto}{Hong Xu}
is a Ph.D. student in the Department of Physics and Astronomy at the University of Southern California. Contact him at hongx@usc.edu.
\end{IEEEbiographynophoto}

\begin{IEEEbiographynophoto}{T.~K. Satish Kumar}
is an artificial intelligence researcher in the Department of Computer Science at the University of Southern California. Contact him at tkskwork@gmail.com.
\end{IEEEbiographynophoto}

\begin{IEEEbiographynophoto}{Nora Ayanian}
is an assistant professor in the Department of Computer Science at the University of Southern California. Contact her at ayanian@usc.edu.
\end{IEEEbiographynophoto}

\begin{IEEEbiographynophoto}{Sven Koenig}
is a professor in the Department of Computer Science at the University of Southern California. Contact him at skoenig@usc.edu.
\end{IEEEbiographynophoto}




\end{document}